\documentclass[journal]{IEEEtran}

\usepackage{lipsum}% http://ctan.org/pkg/lipsum
\usepackage{graphicx}% http://ctan.org/pkg/graphicx
\usepackage{cite}
\usepackage{amsmath,amssymb,amsfonts}
\newtheorem{remark}{Remark}  
\usepackage[linesnumbered,ruled]{algorithm2e}
\usepackage[usenames,dvipsnames,svgnames]{xcolor}
\usepackage{adjustbox}
\usepackage{covington}
\SetKw{KwBy}{by}
\usepackage{caption} 
\usepackage{multirow}
\usepackage{xcolor}
\usepackage{comment}
\usepackage{float}% If comment this, figure moves to Page 2
\usepackage{etoolbox}
\usepackage{scalerel}
\usepackage{tikz}

\usepackage{fancyhdr}

\usetikzlibrary{svg.path}

\definecolor{orcidlogocol}{HTML}{A6CE39}
\tikzset{
  orcidlogo/.pic={
    \fill[orcidlogocol] svg{M256,128c0,70.7-57.3,128-128,128C57.3,256,0,198.7,0,128C0,57.3,57.3,0,128,0C198.7,0,256,57.3,256,128z};
    \fill[white] svg{M86.3,186.2H70.9V79.1h15.4v48.4V186.2z}
                 svg{M108.9,79.1h41.6c39.6,0,57,28.3,57,53.6c0,27.5-21.5,53.6-56.8,53.6h-41.8V79.1z M124.3,172.4h24.5c34.9,0,42.9-26.5,42.9-39.7c0-21.5-13.7-39.7-43.7-39.7h-23.7V172.4z}
                 svg{M88.7,56.8c0,5.5-4.5,10.1-10.1,10.1c-5.6,0-10.1-4.6-10.1-10.1c0-5.6,4.5-10.1,10.1-10.1C84.2,46.7,88.7,51.3,88.7,56.8z};
  }
}

\newcommand\orcidicon[1]{\href{https://orcid.org/#1}{\mbox{\scalerel*{
\begin{tikzpicture}[yscale=-1,transform shape]
\pic{orcidlogo};
\end{tikzpicture}
}{|}}}}

\usepackage[hidelinks]{hyperref}

\ifCLASSINFOpdf

\else

\fi

\begin{document}

\title{Multi-UAV Search and Rescue in Wilderness Using Smart Agent-Based Probability Models 
}

\author{
Zijian Ge \orcidicon{0000-0002-9620-3843}\,,
Jingjing Jiang \orcidicon{0000-0001-7754-9147}\,, \IEEEmembership{ Member, IEEE}, and
Matthew Coombes \orcidicon{0000-0002-4421-9464}\,,\IEEEmembership{ Member, IEEE}
\thanks{Zijian Ge, Jingjing Jiang and Matthew Coombes are with the Department of Aeronautical and Automotive Engineering, Loughborough University, Leicester LE11 3TU, UK, (e-mail:z.ge@lboro.ac.uk; j.jiang2@lboro.ac.uk; m.j.coombes@lboro.ac.uk).} 
}

\maketitle

\begin{abstract}

The application of Multiple Unmanned Aerial Vehicles (Multi-UAV) in Wilderness Search and Rescue (WiSAR) significantly enhances mission success due to their rapid coverage of search areas from high altitudes and their adaptability to complex terrains. This capability is particularly crucial because time is a critical factor in searching for a lost person in the wilderness; as time passes, survival rates decrease and the search area expands. The probability of success in such searches can be further improved if UAVs leverage terrain features to predict the lost person's position. In this paper, we aim to enhance search missions by proposing a smart agent-based probability model that combines Monte Carlo simulations with an agent strategy list, mimicking the behavior of a lost person in the wildness areas. Furthermore, we develop a distributed Multi-UAV receding horizon search strategy with dynamic partitioning, utilizing the generated probability density model as prior information to prioritize locations where the lost person is most likely to be found. Simulated search experiments across different terrains have been conducted to validate the search efficiency of the proposed methods compared to other benchmark methods.

\end{abstract}


\IEEEpeerreviewmaketitle

\section{Introduction}
\IEEEPARstart{W}{ith} the rapid advancement of electronic technologies, including more powerful and affordable onboard computers, improved battery life, advanced sensors, and better communication systems, Unmanned Aerial Vehicles (UAVs) have been widely deployed in various applications such as search and rescue operations \cite{queralta2020collaborative,scherer2015autonomous}, agricultural monitoring \cite{radoglou2020compilation,zhang2021review}, and parcel delivery \cite{wen2022heterogeneous,murray2020multiple}. In particular, Multi-UAV systems equipped with advanced sensors and communication equipment have gained significant attention. The cooperative operation with Multi-UAV can substantially improve mission success rates and work efficiency.

Among various applications, Wilderness Search and Rescue (WiSAR) is particularly well-suited to the solutions provided by Multi-UAV systems. For instance, in 2019 in Queensland, Search and Rescue (SAR) authorities assisted 1,648 people, utilizing 8,733 police person-hours and over 34,000 volunteer hours in SAR \cite{dacey2023understanding}. 
Similarly, the National Missing and Unidentified Persons System in the USA reports that over 600,000 individuals go missing each year, while 70,000 missing persons are reported annually in the UK \cite{lam2023impact}.
%Similarly, between 2000 and 2012, the US National Park Service conducted 4,080 SAR operations at a total cost of 5.3 million, averaging approximately 1,375 per mission \cite{sava2016evaluating}. 
Implementing Multi-UAV systems could significantly reduce both the costs and time associated with these missions compared to using human or canine field teams \cite{wilderness2018technical}.

In WiSAR operations, locating the target quickly is crucial due to the time-sensitive nature of search and rescue efforts. Research indicates that the survival rate of lost hikers and young children aged 4-6 years drops rapidly after the first 24 hours, owing to injuries, exposure, exhaustion, and dehydration \cite{sava2016evaluating}. Long search duration result in a larger area to cover, reducing the likelihood of finding the target. Therefore, minimizing search time not only limits the search area but also maximizes the chances of a successful rescue.

Another significant challenge of WiSAR is operating in rugged terrains such as mountains, forests, and streams, which complicate tracking moving targets. Additionally, weather conditions, as well as the physiological and psychological state of the person, can influence their decisions in wilderness \cite{macwan2014multirobot}. These factors collectively make it difficult to predict such target-specific probability distribution.

Researchers have studied historical search and rescue cases to understand lost individuals' behavior. For instance, \cite{koester2008lost} provides a comprehensive analysis of over 50,000 SAR incidents, detailing factors like recovery locations, survival rates, elevation changes, and dispersion angles. This data helps estimate where a lost person might be. Additionally, regional case studies, such as those in \cite{dacey2023understanding}, \cite{doke2012analysis}, and \cite{pajewski2021predicting}, offer valuable insights for SAR teams before deployment.

Historical datasets have led to various mathematical models for predicting a lost person's location. The Euclidean distance ring model, introduced in 1998 \cite{heth1998characteristics}, uses concentric rings at 25\%, 50\%, and 75\% distances to estimate the probability of the target being within each range. However, it overlooks elevation changes and obstacles. In contrast, the Target Iso-probability Curves from \cite{macwan2011target} incorporate terrain and obstacles, providing a more realistic probabilistic representation. Similarly, a Bayesian approach in \cite{lin2010bayesian} updates prior beliefs about the person's movement based on terrain features using a discretized hexagonal grid map. More recent work \cite{hashimoto2022agent} models complex behaviors by assigning probabilities to actions like seeking higher ground or resting, aiming to better simulate the target’s movement patterns. Despite these advancements, existing methods still neglect how interactions with specific environmental features, such as streams or barriers, impact the probability distribution, which could offer SAR teams more precise location insights.

Multi-UAV search strategies have been extensively researched over the past decades. When dealing with a probability distribution of a lost person over time, the challenge for a Multi-UAV search and rescue team differs from the typical coverage planning problem, which focuses on maximizing area coverage, as discussed in \cite{tian2020search} and \cite{zhu2021multi}. Instead, the emphasis is on searching guided by a probability map. While centralized control approaches, supported by a ground control station, have been employed in studies such as \cite{saxena2019optimal} and \cite{sunehag2017value}, there has been a growing preference for distributed systems due to their fault tolerance, scalability, and adaptability. Notable examples include the distributed real-time search path planning methods using distributed model predictive control introduced in \cite{zheng2023distributed} and \cite{hou2021distributed}, the multi-target search framework proposed in \cite{balanji2024dynamic}, and the moving target search method discussed in \cite{ru2015distributed}. However, many of these methods are based on discretized cell-based systems, which can be limited in resolution and may miss finer details of probability distributions. 

%In terms of information sharing in distributed systems, some work focuses on maintaining communication by ensuring all UAVs remain connected throughout the entire process, as studied in \cite{xu2023cooperative}, \cite{zheng2023distributed}, and \cite{scherer2016persistent}. Alternatively, the recurrent connectivity method, as described in \cite{woosley2020multi}, creates a connected network that only communicates when valuable information is found. 
In distributed systems, continuous communication is often prioritized to keep all UAVs connected, as seen in studies like \cite{xu2023cooperative}, \cite{zheng2023distributed}, and \cite{scherer2016persistent}, focus on keeping all UAVs connected throughout operations, which can ensure stable communication but may limit movement and reduce search efficiency.
An alternative is the recurrent connectivity method \cite{woosley2020multi}, where UAVs communicate only when valuable information is found. However, this approach is designed for exploration and data collection without using a probability map. In WiSAR, where search areas expand 82 $km^2$ every one hour\cite{doherty2014analysis}, commonly used LoRa devices in wilderness settings have a communication range typically limited to 10-20  $km$ \cite{saraereh2020performance}. Thus, maintaining constant connectivity may restrict UAV mobility and efficiency, necessitating a balance between connectivity and search coverage.

In multi-agent search operations, a common strategy is for each UAV to move towards high-probability locations while maintaining a safe distance from other UAVs, as presented in \cite{zheng2023distributed}. This approach works effectively when the probability areas are convex. However, in complex, non-convex probability landscapes, UAVs may converge on the same local high-probability areas, leading to insufficient exploration of unexplored regions.
Some task area assignment methods have been well-investigated for WiSAR to ensure coverage of whole searching area, such as the layered search and rescue algorithm proposed in \cite{alotaibi2019lsar}, which assigns UAVs to each probability layer. Since Iso-probability curves was proposed for predicting a lost person's position by Macwan et al. \cite{macwan2011target}, many researchers have explored specific search strategies utilizing the features of Iso curves. For example, the methods discussed in \cite{macwan2014multirobot} and further in \cite{ku2022wilderness} extend a 3D Iso-probability curve. Similarly, an angular motion control method proposed in \cite{kashino2019multi} and \cite{kashino2020aerial} achieves smooth transitions between curves, also incorporating the concept of a time-related confidence area into the search process. However, traditional task area assignment methods often fall short due to their static nature, where each UAV is assigned a specific zone without the flexibility to adapt as the mission evolves. This rigidity can lead to inefficiencies, particularly in complex environments like WiSAR, where search conditions are highly dynamic.

%The above-discussed methods are optimization-based. Additionally, some geometry-based methods with task allocation have been well-investigated for WiSAR, such as the layered search and rescue algorithm proposed in \cite{alotaibi2019lsar}, which assigns UAVs to each probability layer. Since Iso-probability curves was proposed for predicting a lost person's position by Macwan et al. \cite{macwan2011target}, many researchers have explored specific search strategies utilizing the features of Iso curves. For example, the methods discussed in \cite{macwan2014multirobot} and further in \cite{ku2022wilderness} extend a 3D Iso-probability curve. Similarly, an angular motion control method proposed in \cite{kashino2019multi} and \cite{kashino2020aerial} achieves smooth transitions between curves, also incorporating the concept of a time-related confidence area into the search process. Most of the aforementioned methods assume that the probability distribution is static, and few consider the dynamic environment or search in a discretized map. This can result in the loss of fine details about the terrain, such as minor elevation changes and trails in wildness. In the case of WiSAR, the search strategy needs to be improved to better adapt to moving targets in complex terrains.
\begin{figure}[!t]
    \centering
	\includegraphics[width=0.45\textwidth]{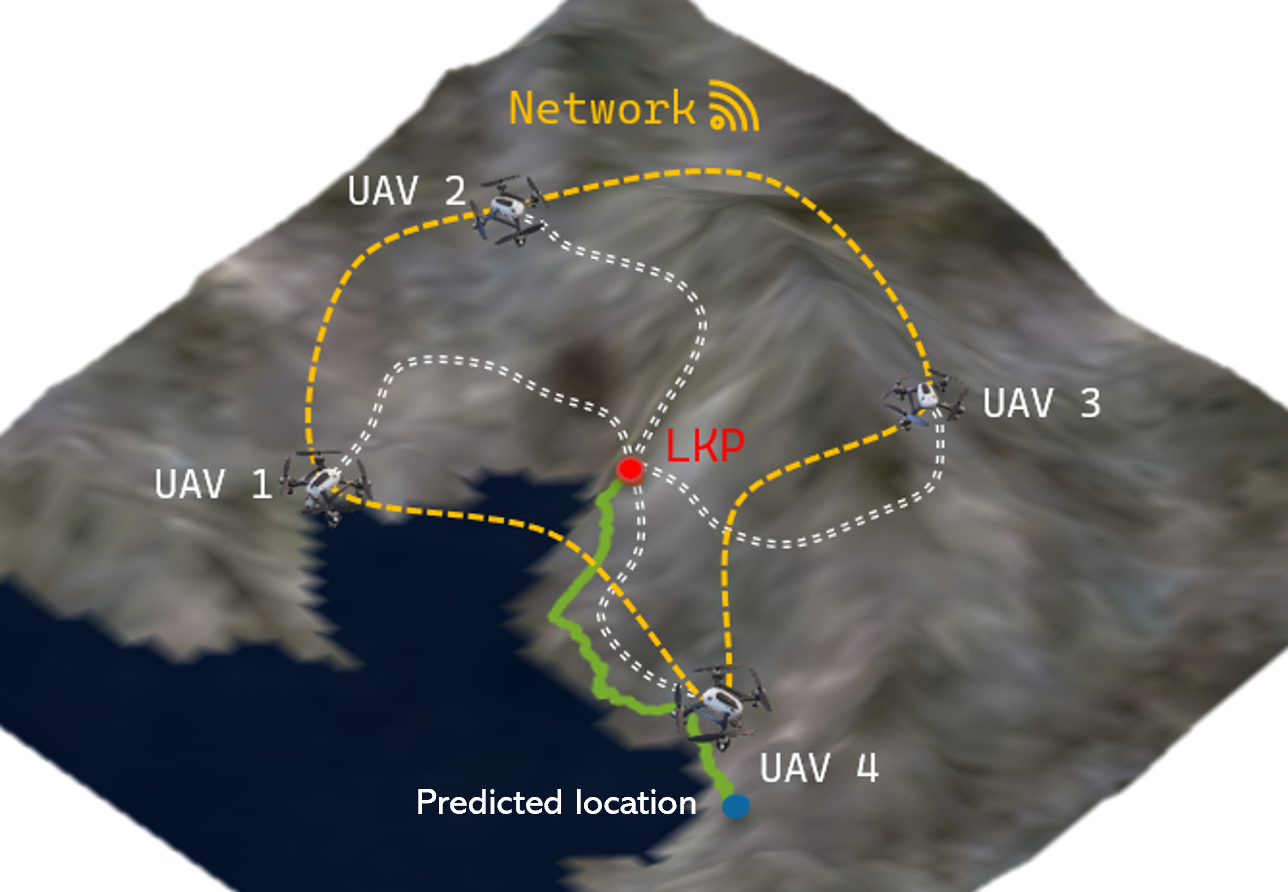}
	\caption{Illustration of a distributed multi-UAV system in WiSAR. UAVs communicate with adjacent UAVs within their communication range. One of the predicted trajectories of the lost person are shown in green line.}
	\label{fig:illustration}
\end{figure}
\begin{figure*}[!t]
    \centering
    \includegraphics[width=0.9\textwidth]{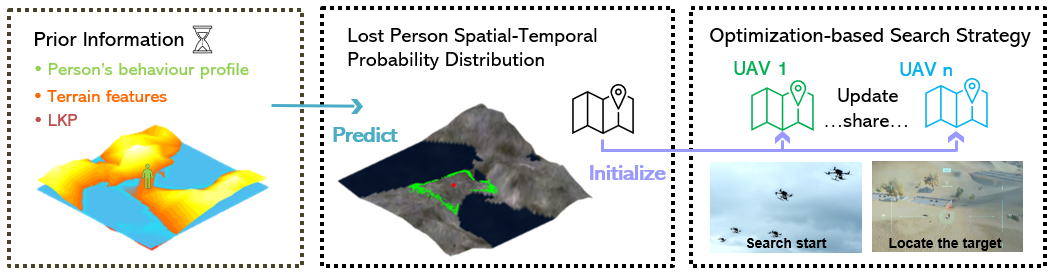}
    \caption{System architecture of the proposed  framework.}
    \label{fig:system}
\end{figure*}

To address above limitations in existing researches, this paper proposes a novel multi-UAV cooperative search strategy designed for a continuous environment, incorporating a topology-based dynamic task area assignment. Additionally, the search is informed by prior information provided through a dynamic probability density map, which is generated using a smart agent-based model. An illustration of the proposed framework work in wildness can be seen in Fig. \ref{fig:illustration}.  The main contributions of this paper are summarized as follows:

\begin{itemize}

    \item  We introduce a smart agent-based probability distribution model, developed using the Monte Carlo method, which integrates terrain features and the lost person's behavior profile. This model simulates decision-making in wilderness settings, such as seeking high-altitude areas for better visibility and following known routes. Additionally, a guideline map is created to ensure that simulated trajectories align with natural paths a person might take, considering obstacles like rivers, streams, and impassable areas.

    \item We establish an optimization-based search strategy that operates in continuous space using a dynamic probability distribution map. A receding horizon mechanism is incorporated into the search process to reduce the likelihood of a single UAV getting stuck in local minima.

    \item A proximity control function is implemented as a soft constraint in the distributed multi-UAV search system. It ensures each UAV maintains an appropriate distance from others, facilitating reliable communication and avoiding collisions in expansive wilderness areas where constant connectivity is challenging. This balance maintains essential communication in the distributed system while minimizing the impact on search coverage.
    
    \item We propose a dynamic partitioning method inspired by Voronoi diagrams for WiSAR missions. This approach assigns each UAV to a specific search area, adapting dynamically as the mission progresses. The method ensures that UAVs do not converge on the same high probability areas and encourages exploration of unsearched regions to maximize coverage.

    %Our approach employs a topology-based dynamic partitioning method to assign UAVs to different probability areas, ensuring complete coverage of the entire search area. For each individual UAV, a receding horizon technique is incorporated into the search process to reduce the likelihood of getting stuck in local minima. 

\end{itemize}

The organization of this paper is as follows: Section \ref{agent model} introduces the smart agent-based probability distribution model, covering the guideline strategy, lost person behavior strategy, and the generation of the dynamic probability map. Section \ref{search strategy} then presents the formulation of the proposed multi-UAV strategy with dynamic task assignment. Finally, Section \ref{simulation analysis} details the simulation results, comparing them with two other benchmark methods, followed by a discussion and conclusions in Section \ref{conclusion}.

\section{Smart Agent-based Model} \label{agent model}
In this section, we describe the prediction of the target's probability distribution, which includes the integration of terrain elevation changes, environmental features, and a profile list into the model. This model is inspired and developed from \cite{hashimoto2022agent} and \cite{macwan2011target}. Figure \ref{fig:system} illustrates the system framework, showing how the estimated probability map assists the multi-UAV rescue team in locating the target.

\subsection{Lost person behavior strategies} \label{behavior list}
Modeling a lost person's potential movement in the wilderness often begins with the Last Known Position (LKP), which refers to the last location where the missing person was confirmed to have been seen or where reliable evidence suggests their presence. The LKP provides a starting point for defining the initial search area. Starting from the LKP, the lost person is modeled as a self-propelled agent moving in discrete time. Their behavior is guided by a probability profile list that includes various possible actions. At each time step, the lost person selects an action to move from the current cell to one of the eight adjacent cells in the grid map. The concept of behavior strategies for a lost person in the wilderness, as defined in \cite{hashimoto2022agent} and based on the research in \cite{koester2008lost}, is extended in our work to continuous space, allowing for more realistic movement simulations and a better reflection of real-world behavior nuances. The main actions of the lost person in the wilderness are defined as follows:

\subsubsection{Random Moving (RM)} The person move randomly in any direction. Which reflect the uncertainties of the lost person's movement, A lost person in the wilderness might move randomly due to disorientation, which can result from unfamiliar terrain, or poor visibility. Panic and stress can further contribute to random movement as the individual may not think clearly and instead wander without a clear direction. 

\subsubsection{Direction Traveling (DT)} This refers to a behavior in which a lost person in the wilderness moves in a consistent, straight-line direction, often guided by a compass bearing or a visible landmark while ignoring trails or paths. This behavior may arise from the belief that following a single direction will lead to safety or from the use of limited navigation tools such as a compass. Environmental factors like the sun or mountain can also influence this directional movement.

\subsubsection{Route Traveling (RT)} This describes a behavior in which a lost person in the wilderness follows a linear feature, such as a trail, road, or ridge line. This behavior often occurs because linear features provide a clear, navigable path that the individual perceives as leading to safety or civilization.

\subsubsection{Staying Put (SP)} This behavior indicates that the lost person stays in the same location rather than moving around. This strategy may involve taking a short break to conserve energy or making visible signals such as large fires or markers to increase the likelihood of being found by rescue teams. 

\subsubsection{View Enhancing (VE)}\label{sec:VE} A lost person attempt to reach a higher vantage point. By climbing to an elevated position, they can see further and more clearly, which helps in spotting rescuers, understanding the terrain, and identifying potential routes or landmarks. 

\subsubsection{Back Tracking (BT)} This involves retracing lost person’s own steps to return to a previously traveled route or landmark. This is also a possible action when the lost person try correct navigation errors.

In this case, all possible actions for a given time step  are defined by the discernment frame $\Lambda$, which contain all aforementioned strategies:
\begin{equation} \label{eq1}
  \Lambda=\{ RM, DT, RT, SP, VE, BT \}, \quad
  \sum_{s \in \Lambda} m(s) = 1
\end{equation}

The weight of a strategy in $\Lambda$ is described by assigning a basic probability \(m\), also referred to as the mass function, to a specific state. For example, \(m(DT) = 0.8\) indicates that the agent has an \( 80\% \) chance of choosing the Direction Traveling strategy at a given time step. The sum of the mass function values derived from $\Lambda$ equals one as presented in Eq. (\ref{eq1}).

\begin{figure*}[]
    \centering
    \includegraphics[width=1\textwidth]{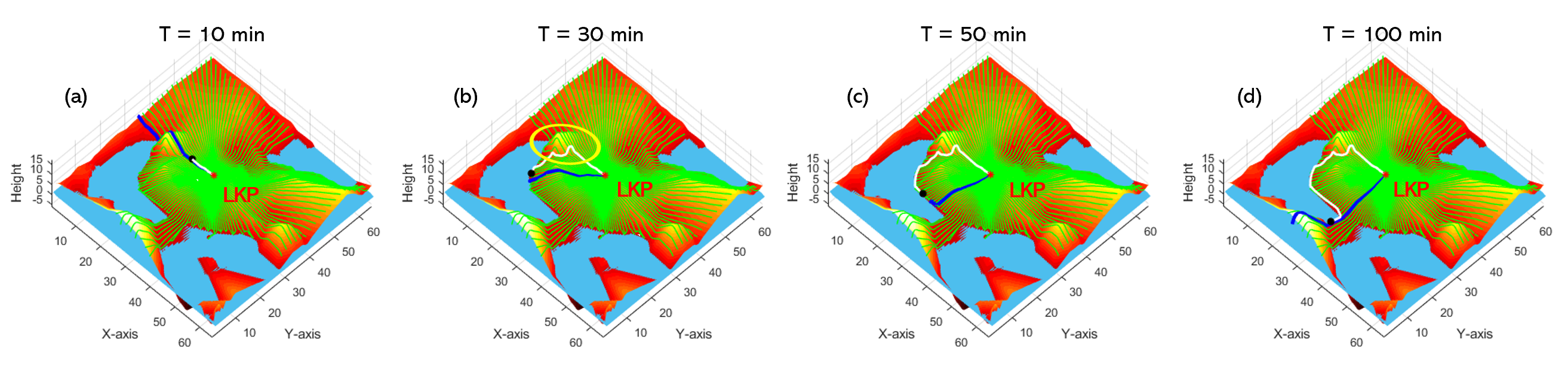}
    \caption{The guideline map is generated using a set of rays originating from LKP, shown as the green lines. A simulated agent (black point) initially selects a ray (blue line) to follow. Upon encountering an uncrossable mountain, the agent switches to an adjacent ray to find a traversable path, following the river's edge until he/she find a way out of the area.}
    \label{fig:Guidelines}
\end{figure*}

\subsection{Guideline Map}
A lost person can move in any direction $\theta \in [0^{\circ},360^{\circ}]$ from his/her LKP. 
A guideline map is developed using a set of reference rays that start from the LKP and are discretized every $n$ degrees (with $n=1^{\circ}$ in our work). Each ray is defined as a linear, obstacle-free path that guides the lost person out of the wilderness area, corresponding to the Direction Traveling strategy explained previously in Section A.

\subsubsection{Ray generation} A single ray is generated based on the idea of the Artificial Potential Field (APF). The ray originating from LKP and extending infinitely in the direction $\theta_j$ determined by the gradient can be written as:
\begin{equation} \label{eq2}
  r_j(t) = r_0 + td_j, \quad  t \geq  0,
\end{equation} 
where $ r_j(t) = (x_j(t), y_j(t))$, denotes the position of any point on the ray at parameter $t$. The LKP is represented by $ r_0 = (x_0, y_0) $ and $d_j$ is influenced by the gradient of the potential field as expressed as follows:
\begin{equation*} \label{eq3}
  d_j = - \nabla U (r_0), \quad  U (r_0) =  U_{rep} (r_0) +  U_{att} (r_0),
\end{equation*} 
where $U_{rep}$ and $ U_{att}$ denote the repulsive potential and attractive potential respectively. The repulsive potential is defined as:
\begin{equation*} \label{eq5}
     U_{rep} = \sum_{i=1} ^{n_o}
\begin{cases}
    \frac{1}{2} k_{rep} (\frac{1}{\lVert r-r_{oi} \rVert} - \frac{1}{d_0} )^2, & \lVert r-r_{oi} \rVert \leq d_0\\
      0, & \lVert r-r_{oi} \rVert > d_0
\end{cases}
\end{equation*} 

The scaling factor of repulsive force is represented by $k_{rep}$, and prior known position of the $i^{th}$ obstacle is expressed by $r_{oi}$ within total number of obstacles $n_o$. $d_0$ is the threshold distance at which the repulsive potential becomes active when the ray is closer to an obstacle, pushing the ray away from it. Similarly, the attractive potential is written as:
\begin{equation*} \label{eq4}
    U_{att} = \frac{1}{2} k_{att} \lVert r-r_{\infty} \rVert ^2,
\end{equation*} 
where $k_{att}$ is the scaling factor of the attractive force, and $r_{\infty}$ is the goal position at infinity in the $\theta$ direction. However, while the ray generated by the APF can navigate small-scale obstacles in the environment, it may fall into a local minimum when encountering large-scale uncrossable area , such as a river. This is a natural limitation inherent in APF methods \cite{pan2021improved}. As a result, the ray may fail to guide the lost person out of the wilderness as depicted in Fig. \ref{fig:Guidelines}. Therefore, a ray transition strategy is proposed in the next section to simulate the behavior of a lost person when dealing with uncrossable areas and rivers.

\subsection{Incorporate the Effect of Terrain}\label{effect of terrain}

The complex terrain topography affects the movement of the target and must be considered when determining the agent model. Two key factors are taken into account: 1) the effect of elevation changes on the lost person's speed, and 2) the presence of uncrossable areas such as rivers, lakes, and steep terrains.

\subsubsection{Terrain map}
In order to consider the terrain difficulty in the smart-agent based model and work in the continuous space, we assume the terrain map is available and can be defined as the probability density function $Terrain(z|x,y)$ which output the elevation value $z$ given the location $(x,y)$. For simplicity, the river or lake areas are identified as those area below a certain elevation $z_0$: 
\begin{equation*} \label{eq5}
   \text{River, Lake Areas} = \{(x,y) \in \mathbb{R}^{n \times 2 }: Terrain(z|x,y) < z_0 \}
\end{equation*}

\subsubsection{Scaled speed}\label{scaled_speed} To incorporate the elevation variation on lost person's motion, a speed scaling strategy is applied for each agent inspired from ISO-probability Curves in \cite{macwan2011target}. Scaling the mean speed of the agent would, in turn, change the distance traveled, and elevation change influence would have to be recomputed for this new scaled travel distance. Assume agent travel from current point $P_{current}$  with a target speed $v_t$ which follow the Gaussian distribution $v_t  \sim \mathcal{N}(v_m,\,\sigma^{2})\, $. With an intended moving direction, the calculated arrival point $P_{next}$ is expressed by: 
\begin{equation} \label{eq5}
   P_{next} = P_{current} +  v_{u} t, \quad v_{u}  \sim \mathcal{N}(v_m q(\gamma),\,\sigma^{2})\,
\end{equation} 
where $v_u$ is scaled speed by the scaling factor $q(\gamma)$ which is calculated through the following steps. 

Step 1: Determine the average slope of the terrain between the current point $P_{current}$ and the intended arrival point $ P_{current} + v_m t $, using the terrain function $Terrain(z|x,y)$ introduced in Section C-(1). The average slope is obtained by fitting a linear regression line to the sample height data points, and can be expressed as $h=ax+b$, and slope value $a$ can be further converted into an angular value $\gamma = \tan^{-1}(a)$.

Step 2: Using the computed average slope, the mean speed $v_m$ of the lost person, adjusted for different terrain slopes, is determined based on empirical data \cite{macwan2011target}. This process results in a linear relationship between the average ground surface slope angle $\gamma$ and the speed scaling factor $q$, which can be expressed as follows:
\begin{comment}
    \begin{equation} \label{eq6}
    q(\gamma) =
\begin{cases}
    q_{dec}(\gamma) = m_1 \gamma + b_1 & \gamma_{min,dec} \leq \gamma < 0^{\circ} \\
      q_{inc}(\gamma) = m_2 \gamma + b_2 & 0^{\circ}  \leq \gamma \leq\gamma_{max,inc}  
\end{cases}
\end{equation} 
\end{comment}
\begin{equation} \label{eq6}
    q(\gamma) =
\begin{cases}
    q_{dec}(\gamma) = 1-\frac{1}{\gamma_{min}} \gamma  & \gamma_{min} \leq \gamma < 0^{\circ} \\
      q_{inc}(\gamma) = 1-\frac{1}{\gamma_{max}}  \gamma  & 0^{\circ}  \leq \gamma \leq\gamma_{max}  
\end{cases}
\end{equation} 
Here, $\gamma_{max}$ and $\gamma_{min}$ represent the maximum incline and maximum decline angles the lost person can handle $(\gamma_{max}>0 , \gamma_{min} <0)$. Note that when the incline angle (decline angle) reaches $\gamma_{max}$ ($\gamma_{min}$), the scaling factor $q_{inc}(\gamma_{max})$ ($q_{dec}(\gamma_{min})$) equals 0, means that agent cannot move forward due to the steep terrain. Additionally, if the surface is flat (\emph{i.e.}, $\gamma = 0^{\circ}$), the scaling factor  $q(0^{\circ}) = 1$. 
%Using these two conditions in a linear equation, the values of the parameters  $m_1,m_2$ and $b_1,b_2$ can be calculated.

\subsubsection{River edge effect}\label{river_effect} A commonly used search strategy by human rescue teams is to follow linear features in the environment, such as streams or river edges, to look for a responsive subject, reserving sweep searches for later efforts \cite{wilderness2018technical}. It makes sense to assume that a lost person encountering a river or stream would follow these nature linear lines in the wildness. 

In our work, a ray transition strategy is proposed within the guideline map to simulate an agent mimicking the behavior of following nature linear lines. At any given time, the agent's position is defined as $P(t)=(x_t,y_t)$, and the agent initially follows a ray $r_i$. A virtual target point $P_t^i$ is assigned on the ray, and the agent moves toward $P_t^i$ based on PID control dynamics. The target point $P_t^i$ is progressively updated along ray $r_i$, ensuring that the agent follow with the ray’s direction.

The agent continuously checks whether the current ray $r_i$ is terminated by impassable terrain. If the ray is terminated, the agent transitions to an adjacent ray. The transition is executed by randomly selecting one of the adjacent rays, $r_{i+1}$ (\emph{i.e.}, right) or $r_{i-1}$ (\emph{i.e.}, left). Once the new ray is chosen, the virtual target point is also updated as $P_t^{i+1}$ or $P_t^{i-1}$. In the case where rays all terminate at the edge of a river, the agent will continuously update its target point, which will align along the river's edge. As a result, the agent will follow the river’s boundary, moving along the river edge. The  trajectory of the simulated agent effected by the river edge can be seen in Fig. \ref{fig:Guidelines}.

\subsubsection{Steep terrain effect} Those impassable steep terrains are different from the prior known environmental obstacles, as they are defined based on the individual's physical ability to navigate steep slopes. If the terrain's slope exceeds the person's maximum incline or decline capacity, it becomes impassable, regardless of whether it might be navigable under normal circumstances explained in Section \ref{scaled_speed}. 

Similarly, the ray transition strategy is applied when the agent encounters impassable terrain, as introduced in Section \ref{river_effect}. In this case, the agent transitions to an adjacent ray if the absolute value of the average slope between the virtual target point \( P_t^i \) and the agent's position \( (x_t, y_t) \) exceeds the maximum allowable angle. As a result, when the simulated agent encounters steep terrain, they will continue switching to adjacent rays until an alternative, navigable route is found, allowing them to resume directional travel. 
For example, as shown in Fig. \ref{fig:Guidelines}-(b), highlighted in the yellow circle, the agent initially attempts to cross a mountain but, upon realizing it is too steep to climb, seeks an alternative route around the terrain.

\begin{figure}[]
    \centering
    \hspace*{-0.16cm}
	\includegraphics[width=0.45\textwidth]{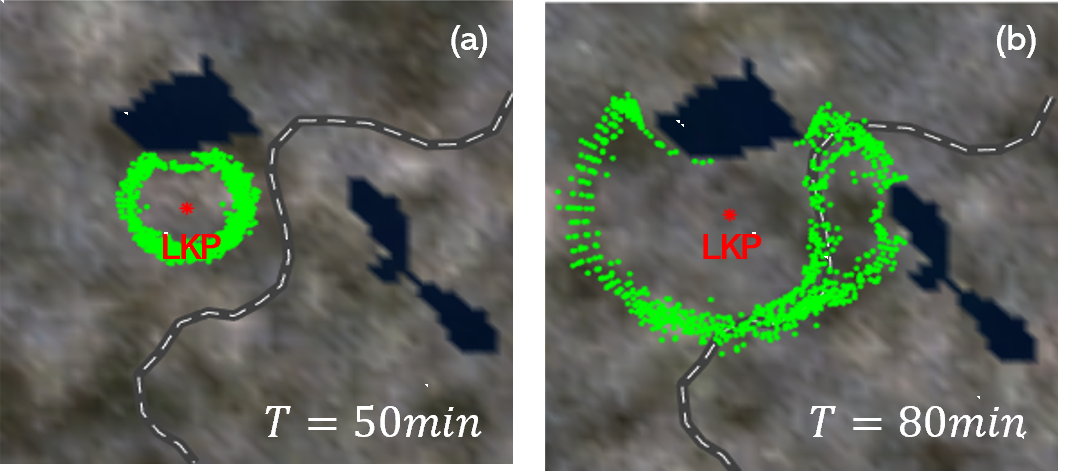}
	\caption{Estimated lost person probability distribution over time: The simulated agents are represented by green particles, and the environmental linear features, such as trails and roads, are shown as the white dash line. }
	\label{fig:particles}
\end{figure}

\subsection{Monte Carlo Simulation}
To estimate the probability distribution of a lost person over time, we use a Monte Carlo simulation with $n_a = 1080$  self-propelled agents, all starting from the LKP. Each agent is assigned an initial reference ray corresponding to his/her travel direction, with movement determined by  behavior strategies list and terrain effects, as outlined in Sections \ref{behavior list} and \ref{effect of terrain}. As the simulation progresses, agents dynamically adjust their paths based on terrain and strategy changes. The resulting spatial distribution of agents represents the likely locations of the lost person, as shown in Fig. \ref{fig:particles}. This distribution then can be further processed into a probability density function to guide UAV search operations by identifying areas with higher search potential. The full procedure for generating the proposed smart agent-based probability model is detailed in Algorithm \ref{smart agent}. 

\begin{algorithm}[]
    \SetKwInOut{Input}{Input}
    \SetKwInOut{Output}{Output}
    \Input{LKP\;
    Terrain Function: $Terrain$, 
    Search duration time: $t_s$\;
    Agent size: $n_a$, 
    Agent speed: $v_a$\;
    Behaviour strategy list: $\Lambda$\; 
     }
    \Output{Predicted agents' spatial-temporal distribution: \\
    $\mathcal{D}$ $\in \mathbb{R}^{n_a \times 2 \times t_s }$;}
    Generate Guidelines:  $Lines = GLine(Terrain, n_a)$\; 

\For{$t=1...t_s$}
      {
       Monte Carlo simulation:\\
       \For{All agents $n=1...n_a$}
       {
       Action selection: $action = Selectaction(\lambda) $ \\
       Scaled speed: $v_{u} = VTerrian(v_a,Terrian,t) $ \\
       Update agent's position: $pos_{n}^t = Move(n,Lines,Terrian,v_{sa}) $ \\
       Save agent's position: $\mathcal{D}$ $\leftarrow$  $pos_{n}^t$ \\
       }
      }  
\caption{Agent-based Probability Model}
\label{smart agent}
\end{algorithm}

%\begin{figure*}[]
   % \centering
    %\includegraphics[width=0.7\textwidth]{dynamic_partitioning_color.png}
    %\caption{dynamic partitioning}
    %\label{fig:dynamic_partitioning}
%\end{figure*}

\section{Distributed Search Strategy} \label{search strategy}

In this section, we begin by introducing the formulation for the multi-UAV search strategy using the smart agent-based model and some techniques we applied to maintain the communication, map update and searching area assignment.

\subsection{Problem Statement}

The main goal of UAV search is to maximize the probability of detecting the target along the search trajectories. In the context of WiSAR, this is defined as the probability of successful target detection ($POS$), which is the product of the probability of the target being in the search area ($POA$) and the probability of the target is detected by the sensor ($POD$), as expressed by:
\begin{equation}\label{eq7}
POS = POA \times POD
\end{equation}
The $POD$ is influenced by factors such as sensor capabilities (e.g., resolution, range, type) and environmental conditions (e.g., weather, terrain, lighting) that affect target visibility. In this work, for simplicity, we disregard sensor failure and environmental conditions, assuming that the target can be detected within the sensor's range, \emph{i.e.}, $POD=1$. Therefore, maximizing $POS$ is equivalent to maximizing $POA$, which is quantified as the cumulative probability within the UAV's Field of View ($FOV$), projected onto a two-dimensional plane as a circular area centered on the UAV's location. In the multi-UAV search problem, each UAV generates a local spatial-temporal probability map using a smart agent-based model. The goal of the multi-UAV search is to explore areas with high probability density while exchanging information and updating the search zones. Each UAV selects an optimal next point to move to within the continuous search space.

\begin{figure}[h]
    \centering
	\includegraphics[width=0.45\textwidth]{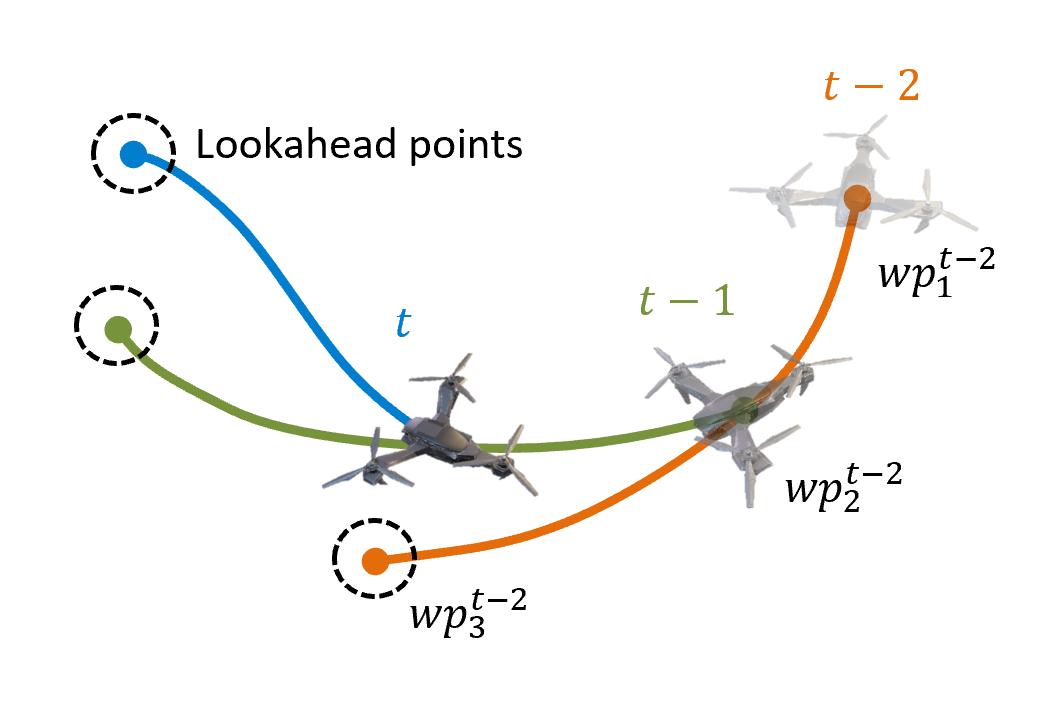}
	\caption{Illustration of the iterations of the receding horizon-based searching.}
	\label{fig:RH_illustration}
\end{figure}

\subsubsection{Objective function}\label{obj fcn}
%Let $wp^c$ denotes the current position of the UAV, and the cost function aims to maximize the value of POA of the next transfer location $wp^{c+1}$, which can be written as:
%Given the current UAV position $wp^t$ at current time $t$, the objective is to find a path made by of a set of waypoints that maximizes the sum of cumulative probability within the search area cover by the path. With a search radius $R$, the path is evaluated over a prediction horizon of length $l$. The optimization problem can be formulated as:

Given the current UAV position $wp^{t}_1$ at time $t$, the objective in each iteration is to find a path $wp^{t}$ consisting of a set of waypoints (number of waypoints is defined by $n_l$) that maximizes the cumulative probability of detecting the target within the area covered by the path.  The optimization problem can be formulated as:
%\begin{equation}\label{eq8}
%\begin{aligned}
%wp^{c+l} &= \arg\max_{x,y} POA(P^t(x,y)) \\
%&=  \int_{\substack{(x, y) \in \text{area}(wp^c, wp^{c+l}, R)}} P^t(x, y) %\, dx \, dy,
%\end{aligned}
%\end{equation}
\begin{equation}\label{eq8}
wp^{t} =  \arg\max_{wp^{t}_1 \rightarrow wp^{t}_{n_l}}  \sum_{j=1} ^{n_l} \alpha POA_j(wp_j,R),
\end{equation}
%where $POA_j(wp^j, R)$ represents the cumulative probability of detection within the search radius $R$, centered at waypoint $wp_j$, based on the probability density function generated by the smart agent-based model (as explained in Remark \ref{rm1}). The path $wp_1^t \rightarrow wp_{2}^t \rightarrow \cdots \rightarrow wp_{n_l}^t$ is considered over the prediction horizon, but only $wp^{t}_2$ is chosen as the next position ($wp^{t+1}_1 = wp^{t}_2$) for the UAV to transfer. It is important to notice that the length between current location to next location is determined by product of the predefined UAV flying speed $v_u$ and time interval $\Delta t$ as $\lVert wp_1^t - wp_2^t \rVert = v_u  \Delta t$. While the length between other waypoints is defined as a tune-able parameter which defined as the length of the prediction horizon $l$, which can be written as $l = \lVert wp_i^t - wp_{i+1}^t \rVert$ $(1<i<n_l-1) $.
where $POA_j(wp^j, R)$ represents the cumulative probability of detection within the search radius $R$, centered at waypoint $wp_j$, based on the probability density function generated by the smart agent-based model (as explained in Remark \ref{rm1}). $\alpha$ is the scaling factor which satisfies $\alpha > 1$. The path  $wp_1^t \rightarrow wp_{2}^t \rightarrow \cdots \rightarrow wp_{n_l}^t$ is considered over the prediction horizon with horizon length $n_l$, but only $wp^{t}_2$ is executed as the next position for the UAV, meaning $wp^{t+1}_1 = wp^{t}_2$. 

%It is important to note that the distance between the current location and the next waypoint is determined by the product of the predefined UAV flying speed $v_u$ and the time interval $\Delta t$, as $\lVert wp_1^t - wp_2^t \rVert = v_u  \Delta t$. 

%The distance between the current location and the next waypoint is determined by the UAV's flying speed $v_u$ and the time interval $\Delta t$, expressed as $\lVert wp_1^t - wp_2^t \rVert = v_u \Delta t$. For consecutive waypoints along the planning horizon, the distance between each pair, referred to as the step size, is a tunable parameter expressed as $d_p = \lVert wp_i^t - wp_{i+1}^t \rVert$ $(1 < i < n_l - 1)$.

For consecutive waypoints along the planning horizon, the distance between each pair, referred to as the step size, is determined by the UAV's flying speed $v_u$ and the time interval $\Delta t$, expressed as $\lVert wp_1^t - wp_2^t \rVert = v_u \Delta t$. The horizon length, $n_l$, is a tunable parameter. A larger $n_l$ generates more waypoints further ahead, enabling more optimal and informed decision-making. However, this also increases the computational load.

%this comes at the cost of making the UAV less reactive to immediate changes, as it prioritizes a longer-term path, and also increases the computational load.

An illustration of the receding horizon search mechanism can be seen in Fig. \ref{fig:RH_illustration}. 
Introducing a planning horizon in UAV search operations by sampling potential future trajectories (defined as lookahead points in this work) allows UAVs to effectively explore complex, dynamic environments and avoid livelock or getting stuck \cite{zhang2014recursive}, especially in our non-convex probability distribution map.
%Adding a prediction horizon in UAV search operations helps UAV to evaluate short-term paths and avoid local minima, particularly in dynamic environments with non-convex probability distributions. 

\begin{remark}\label{rm1}
The smart agent-based model provides the spatial distribution of particles, where a higher concentration of particles in a given area indicates a higher probability of the target being present in that area. However, to integrate this into path planning, we must further process the particle distribution into a probability density function using Kernel Density Estimation (KDE), as demonstrated in our previous work in \cite{ge2023congestion}. The spatial-temporal density function is represented by:
\begin{equation} \label{density}
	\hat{f}_s (x,y,t) =\frac{1}{n h^2_s} \sum_{i=1}^{n} K_s(\frac{x-x_i^t}{h_s},\frac{y-y_i^t}{h_s}), \
\end{equation}
where \((x_i^t,y_i^t)\) represents the location of particle $i$ at time $t$, and $n$ is the total number of particles. \(\hat{f}_s (x,y,t)\) is the estimated density value at position \((x,y)\), with  \(h_s\) as the spatial bandwidth and \(K_s\) as the kernel function. A detailed evaluation of using KDE to generate the density map can be found in \cite{ge2023congestion}.
\end{remark}

\subsubsection{Turning angle constraint} \label{constraint:turning angle}
The turning angle refers to the horizontal angle formed between the UAV's current direction and its previous trajectory. For smooth and practical maneuvering  without sharp turns, the UAV’s path must maintain a turning angle below a certain threshold. Given the current position $wp^t_1 = (x_i,y_i)$, the constraint can be written as:
\begin{equation}\label{max_angle}
\left| \arctan\left(\frac{y_{i+1} - y_i}{x_{i+1} - x_i}\right) - \arctan\left(\frac{y_i - y_{i-1}}{x_i - x_{i-1}}\right) \right| \leq \theta_{\text{max}},
\end{equation}
where $\theta_{max}$ is the maximum turning angle that UAV can handle.

\begin{figure*}[ht]
    \centering
    \includegraphics[width=0.75\textwidth]{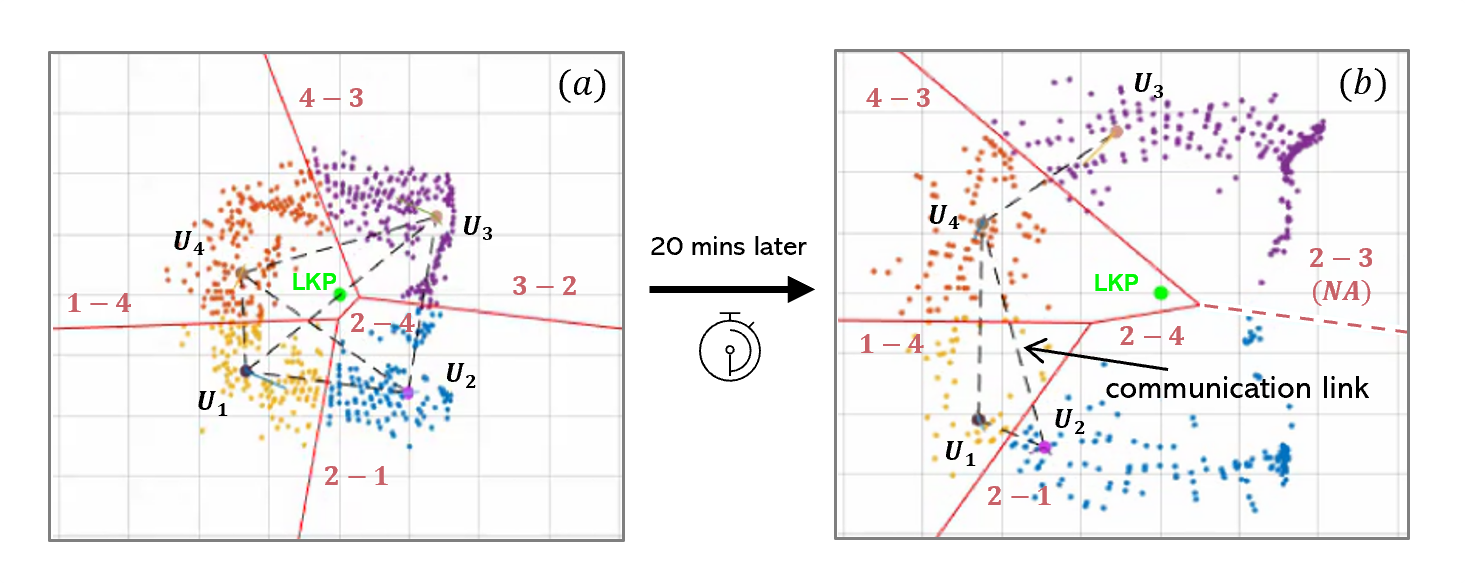}
    \caption{Multi-UAV search with dynamic probability map using Voronoi partitioning: Different UAVs are assigned to distinct probability areas, represented by different colors. The solid red line indicates an active boundary of a Voronoi cell, which is formed when two UAVs are within communication range. In (b), the dashed red line shows a virtual partition between UAV 3 and UAV 2 that does not exist, as there is no connection between them.}
    \label{fig:dynamic_partitioning}
\end{figure*}

\subsubsection{Distance constraint} \label{constraint:distance}
In our multi-UAV search for WiSAR, maintaining effective network connectivity is crucial for information sharing and coordination. However, as discussed in the introduction, to avoid unnecessary restrictions on UAV movement, our distance constraints are only activated between each UAV and its nearest adjacent UAVs, rather than across the entire fleet. This approach is particularly suited to the ring-shaped probability area, as shown in Fig. \ref{fig:particles}, where the search radius expands over time. As UAVs move outward from the center, maintaining distance constraints with nearby UAVs allows for greater flexibility in movement. This design prevents UAVs from being constrained to stay close to the entire group, enabling them to fully explore the expanding search area. In our work, we apply a soft constraint for maintaining connectivity using a proximity function.

Let $U_i$ $\forall i \in \{1,2,...,n\}$ represent the set of $n$ UAVs in the search team, and $f(d_{ij}^t)$ denotes the proximity function between adjacent $U_i$ and $U_j$, which can be expressed as:
\begin{equation}\label{eq9}
f(d_{ij}^t) =  \frac{k_1}{(d_{ij}-d_{min})^ {\xi}} + \frac{k_2}{(d_{max}-d_{ij})^ {\xi}},
\end{equation}
where, $d_{ij}^t$ is the Euclidean distance between two adjacent UAVs, $U_i$ and $U_j$, at time $t$, as calculated by  $\lVert wp^{t,i} -  wp^{t,j} \rVert$, $d_{min}$ is the minimal distance allowed between UAVs to avoid collision and $d_{max}$ is defined as the maximum communication range.  $k_1$, $k_2$ are the scale factor that adjust how quickly the function grows near $d_{min}$ and $d_{max}$, and $\xi$ is exponent factor that control the steepness of the proximity function. The higher the exponent, the more sharply the function grows near the boundaries. 

Let $V={1,2,...,n_c}$ be the set of nodes (UAVs) in the network and $E_{c}$ as the set of edges between each $U_i$ and its closest adjacent $U_j$, which can be determined as:
\begin{equation*}\label{eq10}
j = \arg\min_{k \in V, K \neq i} d_{ik}
\end{equation*}
Subsequently, the soft constraint for collision avoidance and communication maintenance can be defined by minimizing the proximity control function for each UAV with respect to its nearest UAV, as expressed by:
\begin{equation*}\label{eq11}
\min_{wp^{t,1},wp^{t,2},...,wp^{t,n_c}}  \sum_{i=1} ^{n_c} \epsilon f(d_{ij}^t)
\end{equation*}
Here, $\epsilon$ is a scaling factor. Notably, if UAVs are either too close or too far from adjacent UAV, they incur a penalty from this function, particularly when approaching the boundary values ($d_{min}$ and $d_{max}$), where the proximity function increases exponentially. However, this does not significantly affect the UAVs' motion flexibility in exploring high-probability areas, as the impact of this term, with $0< \epsilon < 1$, is relatively small compared to the scaling factor $\alpha$ in the main objective function in Eq. \ref{eq8}.

\subsubsection{Cognitive map update for dynamic environments}
In our multi-UAV search strategy, each UAV is initially provided with a spatial-temporal distribution of all simulated agents generated by the smart-agent-based model. This distribution, denoted as \( \mathcal{D}_i(t) \) for each \( U_i \), reflects the estimated locations of all \( n_a \) agents over time. As the agents move, the search distribution is dynamically updated to reflect changes in the environment.

If two UAVs come within communication range, they exchange information about which agents have been searched, using a pheromone-based mechanism. Specifically, when \( U_i \) detects an agent \( j \), it leaves a pheromone marker \( \phi_{ij}(t) \) on the detected agent \( j \) at time \( t \). This marker indicates that the agent has been searched, and this information is shared with other UAVs within the communication range \( R_c \). The update of the pheromone information can be described as:
\[
\phi_{ij}(t) = 
\begin{cases}
1, & \text{if agent } j \text{ is detected by $U_i$ } \text{ at time } t \\
0, & \text{otherwise}
\end{cases}
\]

Each UAV within the communication range \( R_c \) of \( U_i \) will receive the pheromone information \( \phi_{ij}(t) \). As a result, the distribution map of \( U_i \) will be updated to exclude those agents that have been marked as searched:
\[
\mathcal{D}_i(t+1) = \mathcal{D}_i(t) - \{ j \mid \phi_{ij}(t) = 1 \}
\]
where \( \mathcal{D}_i(t+1) \) represents the updated distribution map for \( U_i \) after receiving pheromone information. This ensures that \( U_i \) no longer considers agents \( j \) that have already been searched.

\subsubsection{Dynamic partitioning} \label{constraint:dynamic partioning}
The search area is partitioned among the connected UAVs using the Voronoi partitioning method, as inspired by the works in \cite{hu2020voronoi} and \cite{nanavati2024distributed}. The centroid of each Voronoi cell corresponds to the position of a UAV, ensuring that each UAV is allocated a specific region to search. As the UAVs move, the Voronoi partitions dynamically adjust based on the updated UAV positions and the available communication network.

%Lets define the Voronoi cell by a convex polygon $Q \subset \mathbb{R}^2$. $R_i$ $\forall i \in {1,2,...,n}$ are connected UAVs and $N_{Ri}$ is a UAV set satisfying that each element in this set is a neighbor UAV of $R_i$, and $p_i$ represents the position of robot $R_i$. The Voronoi cell $Vor(R_i)$ is defined by:

%Let $U_i$ $\forall i \in \{1,2,...,n\}$ represent the set of $n$ connected UAVs, and 
Let $N_{U_i}$ be the set of neighboring UAVs within communication range for $U_i$. The position of $U_i$ is given by $p_i$. The Voronoi cell Vor$(U_i)$, representing the region controlled by $U_i$, is defined as the set of all points closer to  $U_i$ than to any other UAV  $U_j$, $j \neq i$. Mathematically, it can be expressed as:
%corresponding to the $ith$ UAV denoted by $V_i$, based on its network neighbours at any time instant $t$. The Voronoi cell can be expressed mathematically by:
%V_i = \{ q \in D \mid \lVert p_t^{(i)} - q \rVert \leq \lVert p_t^{(j)} - q \rVert \quad \forall j \neq N_{c,t}^{(i)}   \setminus i   \},
\begin{equation}\label{voronoi}
\text{Vor}(U_i) = \{q \in Q \mid \|q - p_i\| \leq \|q - p_j\|, \, \forall U_j \in N_{U_i}\},
\end{equation}
where $Q$ is the Voronoi partition existed as a convex polygon, generated by $U_i$. To integrate this partitioning strategy into the search problem, each $U_i$, located at position $p_i$ at time instant $t$, only needs to know the boundaries of its own region $Q$. These boundaries consist of linear bisectors, which are the perpendicular lines between $U_i$ and its neighboring connected UAVs. For example, as shown in Fig. \ref{fig:dynamic_partitioning}, the search area $Q$ allocated to $U_1$ is bounded by the perpendicular bisectors between $U_1$ and $U_4$, as well as between $U_1$ and $U_2$, denoted as lines $1-4$ and $2-1$, respectively. Further then the constraint for the $U_i$ with its position $p_i=(x_i,y_i)$ and any connected $U_j$ with $p_j=(x_j,y_j)$ at time $t$, which forms part of the Vornoi cell can be written as:
\begin{align*}\label{eq:pareto mle2}
y_i < a_j^tx_i + b_j^t,  j \in N_{U_i} \quad \text{if ego $U_i$  below the line}\\
y_i > a_j^tx_i + b_j^t,  j \in N_{U_i} \quad \text{if ego $U_i$ above the line}
\end{align*}
The perpendicular bisector, expressed as  $y = a_j^tx_i + b_j^t$, is dynamically determined by the positions of two connected UAVs. The insights of using dynamic partitioning is explained in Remark \ref{rm2}.
\begin{remark}\label{rm2}
In our work, dynamic partitioning is applied for handling non-convex and dynamic probability maps in multi-UAV search operations. This approach effectively prevents UAVs from clustering around the same high-probability areas, promoting exploration of previously unsearched regions by assigning different search zones to each UAV.
\end{remark}
%In our work, dynamic partitioning is applied for handling non-convex and dynamic probability maps in multi-UAV search operations. This approach effectively prevents UAVs from clustering around the same high-probability areas, promoting exploration of previously unsearched regions by assigning different search zones to each UAV.

\subsection{Problem Solving}
The proposed search strategy is formulated as an optimization problem with an objective function defined in Section \ref{obj fcn} and constraints described in Sections \ref{constraint:turning angle} to \ref{constraint:dynamic partioning}. %This optimization problem can then be solved using an appropriate optimization solver. 
Since the development of a nonlinear solver is not our focus, we use CasADi \cite{Andersson2019} in our work, a powerful software framework that offers flexibility in modeling and solving complex nonlinear problems. 
%Finally the overall search algorithm of each UAV in the team is summarized in Algorithm \ref{algorithm2}. 

\begin{comment} 
\begin{algorithm}[t] 
    \SetKwInOut{Input}{Input}
    \SetKwInOut{Output}{Output}
    \Input{Search start location LKP\;
    Available search time: $t_s$, and pre-set flying speed: $v_u$\;
    Agent-based probability model: $\hat{f}_s (x,y,t)$ \;
    Number of UAVs involved for searching : $n_u$\;
    {lookahead step: $n_l$}\;  
     }
     \Output{Reference route points: $Path$;}
    \textbf{Initialize}: Assign initial $\hat{f}_s (x,y,t)$ to each UAV's cognition map $\hat{f}_s (x,y,t)^k$, $k=1...n_u$ \;
\For{$t=1...t_s$}
      {
      Create variables: $ wp^t \in \mathbb{R}^{n_{l} \times 2}$ 
      
      \If{reach the search area}{
       
      Information exchange with linked UAVs to update $\hat{f}_s (x,y,t)^k$\;
       Solve formulated optimization problem to obtain  $wp^{t}_2$\;
       Route update:
       $Path$ $\leftarrow$ add $wp^{t}_2$ \;
       
      }

      }  
\caption{Multi-UAV Search Strategy}
\label{algorithm2}
\end{algorithm}

\end{comment}

\section{simulation analysis} \label{simulation analysis}
In this section, simulated experiments with two comparison methods were performed to validate the proposed Multi-UAVs search strategy with smart agent-based probability model. All the simulations are conducted on a desktop computer (Intel i5-11500 processor, 16GB RAM).

\subsection{Mission Scenarios and Parameter settings}

% \usepackage{multirow}
\begin{table*}[]
\caption{Comparison of different search methods based on search time and success rate.}
\centering
\scalebox{1.1}{
\begin{tabular}{cccc|cc|cc}
\cline{3-8}
                                         &                          & \multicolumn{1}{c|}{Success rate} & Average time & \multicolumn{1}{c|}{Success rate} & Average time & \multicolumn{1}{c|}{Success rate} & Average time \\ \hline
\multicolumn{2}{c|}{Terrain roughness}                              & \multicolumn{2}{c|}{Mild roughness}              & \multicolumn{2}{c|}{Moderate   roughness}        & \multicolumn{2}{c}{Severe  roughness}           \\ \hline
\multicolumn{2}{c|}{RHS}                                            & \multicolumn{1}{c|}{46\%}         & 57.65 min    & \multicolumn{1}{c|}{50\%}         & 52.96 min    & \multicolumn{1}{c|}{58\%}         & 40.85 min    \\ \hline
\multicolumn{2}{c|}{ISO}                                            & \multicolumn{1}{c|}{42\%}         & 78.03 min    & \multicolumn{1}{c|}{36\%}         & 81.61 min    & \multicolumn{1}{c|}{32\%}         & 79.52 min    \\ \hline
\multicolumn{2}{c|}{TPS}                                            & \multicolumn{1}{c|}{40\%}         & 63.70 min    & \multicolumn{1}{c|}{48\%}         & 61.77 min    & \multicolumn{1}{c|}{52\%}         & 60.21 min    \\ \hline
\multicolumn{1}{c|}{\multirow{2}{*}{$T_{e}$,$V_{e}$}} & \multicolumn{1}{c|}{ISO} & \multicolumn{1}{c|}{9.52\%}       & 26.12\%      & \multicolumn{1}{c|}{38.89\%}      & 35.11\%      & \multicolumn{1}{c|}{81.25\%}      & 48.63\%      \\ \cline{2-8} 
\multicolumn{1}{c|}{}                    & \multicolumn{1}{c|}{TPS} & \multicolumn{1}{c|}{15.00\%}      & 9.50\%       & \multicolumn{1}{c|}{4.17\%}       & 14.26\%      & \multicolumn{1}{c|}{11.54\%}      & 32.15\%      \\ \hline
\end{tabular}
}
\label{table:result}
\end{table*}

\subsubsection{Terrain Generation} The terrain map used in the simulation is generated through a MATLAB package: Automatic Terrain Generation 

\cite{terrain_generation}. The tool can generate a realistic-looking terrain on a two dimensional surface described by a  discretized terrain matrix, which can be further use to approximate the terrain function through normalization and probability interpretation technique. In addition, to evaluate the proposed search strategy in environment with different terrain complexity, the key parameters for adjusting the terrain map are provided as follow:
\begin{equation*}\label{eq:terrain}
\text{Terrain Parameters Matrix: } [n_e,el,r_0,r_r],
\end{equation*}
where $n_e$ represents the size of the terrain map $n_e \times n_e$, and $el$ is the initial elevation. Increasing $el$ raises the overall altitude of the terrain. $r_o$ indicates the initial roughness, with higher values producing more extreme, jagged variations, such as larger peaks, valleys, and a higher likelihood of rivers and lakes. Finally, $r_r$ is the roughness rate, which determines how roughness changes over distance. A high $r_r$ causes the terrain to smooth out more quickly, while a low $r_r$ maintains roughness over a larger area, preserving the terrain’s variations.

\subsubsection{Lost person model} For evaluation of different search frameworks. We assume that the search is ordered to locate a lost person. The target is set to be an experienced hacker which follow the behavior model in \cite{hashimoto2022agent}. The walking speed is represented by a normal distribution with a mean of $\mu = 0.75m/s$ and a standard deviation of $\sigma =0.25m/s$, suggested by the research in \cite{koester2008lost}.

\begin{figure}[t]
    \centering
	\includegraphics[width=0.4\textwidth]{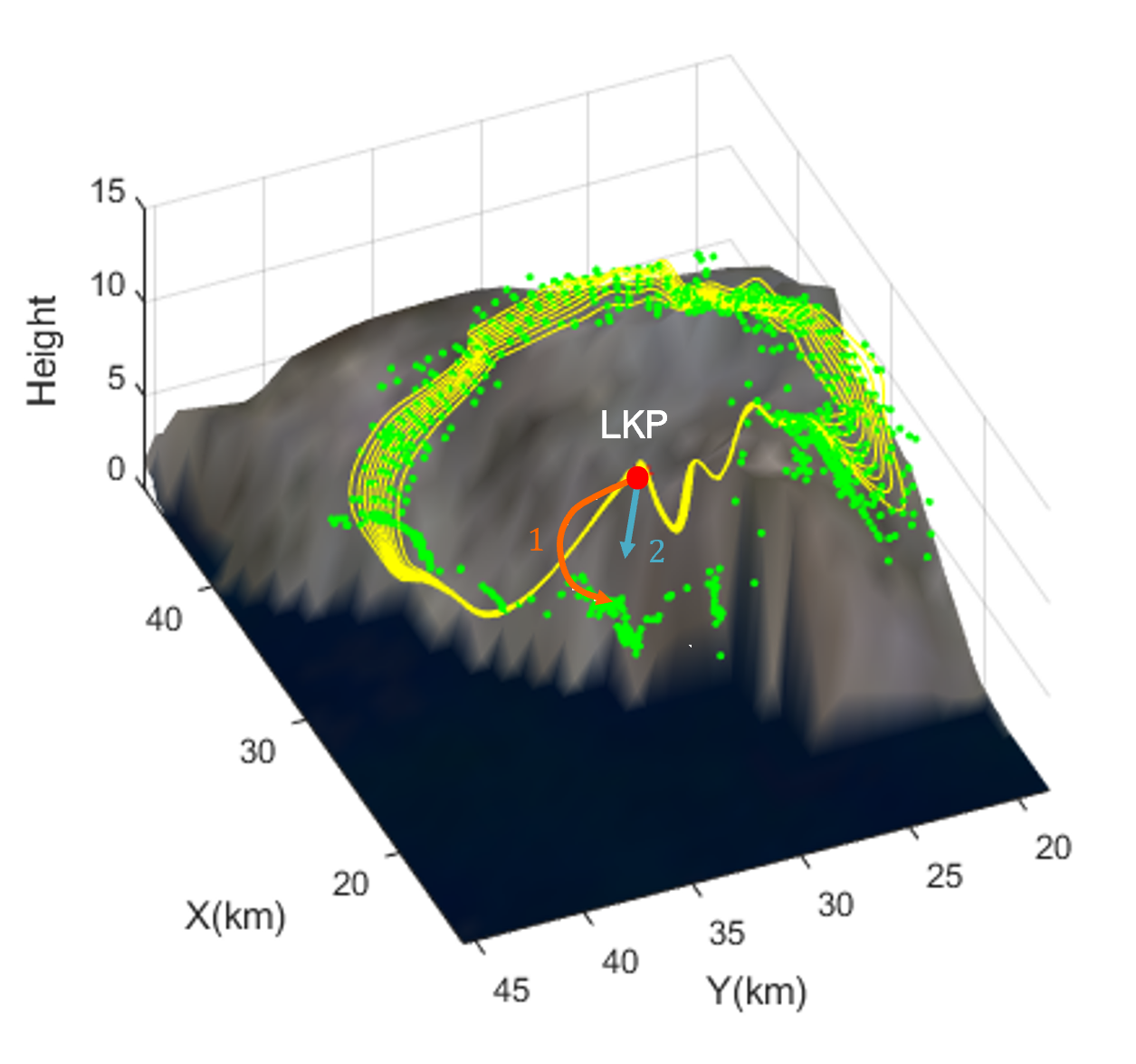}
	\caption{ A illustrative comparison of ISO-probability curve and our smart agent-based model to estimate the lost person probability distribution. }
	\label{fig:iso_illustration}
\end{figure}

\subsubsection{Mission briefing}
In this simulation setup, the search area spans  $65km \times 65km$, with the target starting at the LKP located at the center, $[30km,30km]$. A team of four UAVs is deployed, each flying at a speed of $5 m/s$, and they begin the search 30 minutes after the lost person is reported from the LKP. The UAVs are equipped with LoRa communication devices, giving them a maximum communication range of $15 km$. The FOV for each UAV is modeled as a circle with a radius of 50m. Altitude changes are not considered in this simulation, meaning the FOV is assumed to project consistently onto the terrain map without variation due to elevation. The total search time is limited to 2 hours. If the UAV team fails to locate the target within this time, the mission is deemed unsuccessful.

\begin{figure*}[t]
    \centering
	\includegraphics[width=1\textwidth]{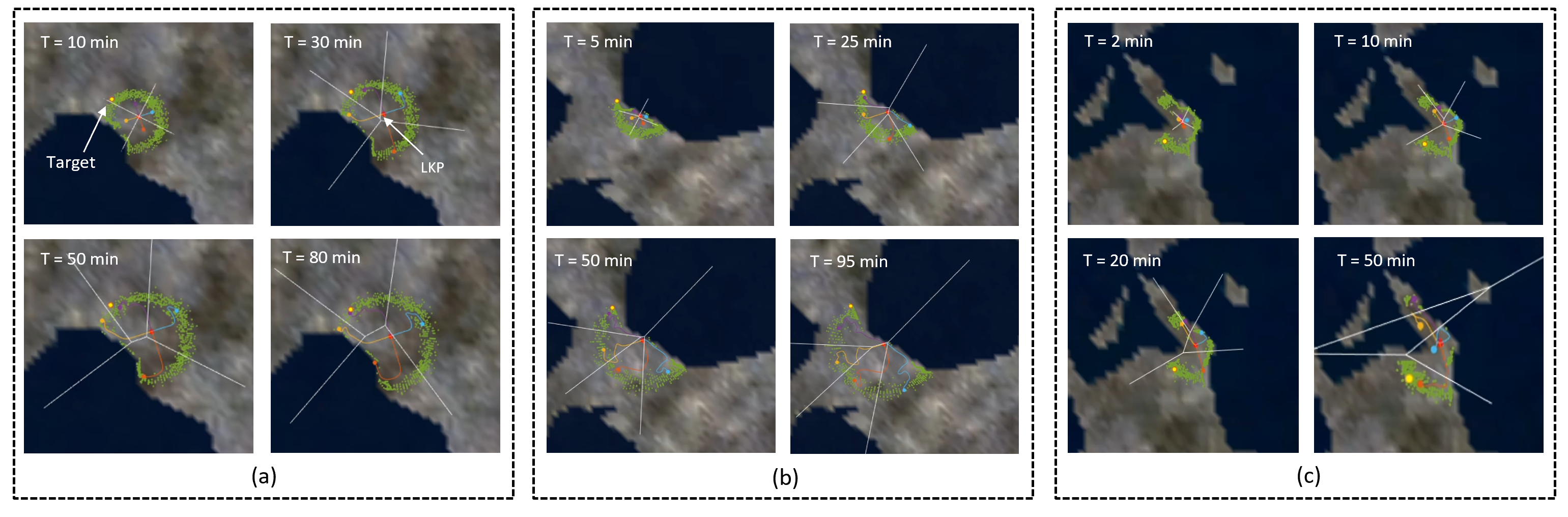}
	\caption{ Simulation examples show the trajectories of the proposed Multi-UAV search method with smart agent-based model across different terrain maps. The maps illustrate varying levels of terrain roughness: Mild (a), Moderate (b), and Severe (d). }
	\label{fig:results}
\end{figure*}

\subsection{Evaluation Method}
To evaluate the proposed search strategy with the smart agent-based model, two benchmark search frameworks are selected for comparative studies. The first benchmark is a distributed multi-UAV search method using the common formulation for multi-agent systems without a dynamic partitioning mechanism, as proposed in \cite{zheng2023distributed}. In this case, we replace their reference probability map with our smart agent-based probability map. The second method is a widely used geometry-based search approach for WiSAR, developed using ISO-probability curves, as detailed in \cite{kashino2019multi}.

\subsubsection{Evaluation index}To quantify the improvement in search efficiency of the proposed method compared to the benchmark methods,  evaluation indexes are defined as:
\begin{equation} \label{evaluation}
T_{e}   = \frac{T_{ben}-T_{s}}{T_{ben}}, \quad V_{e}   = \frac{V_{s} - V_{ben}}{V_{ben}}, 
\end{equation}
 where \(T_{s}\) and \(T_{ben}\) denote the average search time for the proposed method and the benchmark method, respectively. Similarly, \(V_s\) and \(V_{ben}\) represent the mission success rate of the proposed method and the benchmark method, respectively. Furthermore, \(V_e\) and \(T_e\) indicate the improved search efficiency and the improved success rate  compared to the benchmark method.

\subsection{Reference Model Analysis}
The ISO-probability curve, proposed in \cite{macwan2011target} and used in one of our benchmark methods \cite{kashino2019multi}, is a widely used approach for estimating the probability distribution of a lost person's location over time, while accounting for terrain. The core concept involves generating sets of probability bounds (e.g., $ 10\%, 20\%,...,90\% $), with each bound corresponding to a specific probability curve. These curves represent areas where the cumulative probability of finding the lost person within the boundary is specified. For instance, the $50\%$ curve indicates a $50\%$ likelihood that the lost person is within the enclosed area. 

While historical data for analyzing numbers of lost persons in specific locations and terrains is unavailable, making theoretical proof of the accuracy of our model and the ISO-probability curves impossible, there are notable differences observed in simulations. In terrains without complex features such as rivers or steep slopes, both models perform similarly since they both account for elevation changes affected by target's speed. However, our model, which incorporates the lost person’s behavior strategies, offers more detailed and nuanced probabilities. For example, local high areas may attract more particles in our model, reflecting the "View Enhancing" strategy may executed by the agents, as described in Section \ref{sec:VE}. Consequently, our model results in higher granularity and detail within the bounds of the ISO-probability curves, while maintaining a consistent overall probability distribution across the ISO bounds. 
%although the overall probability distribution within the ISO bounds. 
A key distinction arises in terrain maps with complex environmental features. In such cases, the self-propelled smart agents in our model interact directly with these features. For instance, as illustrated in Fig. \ref{fig:iso_illustration}, when starting from an LKP located on a hillside, ISO-probability curves may halt expansion due to the terrain being too steep, as identified by the control points. However, in reality, a lost person may attempt to find an alternative route. In our model, agents do just that—detouring to find a travelable path. For example, in Fig. \ref{fig:iso_illustration}, an agent originally moving in the blue-arrow direction encounters a steep area and finds an alternative route (orange path). Additionally, effects such as river boundaries, as explained in Section \ref{effect of terrain}, further differentiate our probability estimation from that of the ISO-probability curves.

\subsection{Simulation Results}

\begin{comment}
\begin{figure}[]
    \centering
    \hspace*{-0.15cm}
	\includegraphics[width=0.45\textwidth]{results1.png}
	\caption{A simulation example of the proposed Multi-UAVs search method with moderate terrain roughness. Search at 10 min (a), 30 min (b), 50 min (c), 80 min (d).}
	\label{fig:results1}
\end{figure}
\end{comment}

To evaluate the effectiveness of our multi-UAV search strategy in Wildness, we selected three different terrain complexities, each generated using different terrain parameters as seen in Table \ref{table:parameters}. 

\begin{table}[h]
\caption{Parameters setting of different terrain roughness}
\centering
\scalebox{1.1}{
\begin{tabular}{l|l|l|l}
\hline
Terrain   Roughness & $el$ & $r_0$ & $r_r$ \\ \hline
Mild                & 8  & 6  & 10  \\ \hline
Moderate            & 8  & 11  & 18  \\ \hline
Severe              & 5  & 18  &  8  \\ \hline
\end{tabular}
}
\label{table:parameters}
\end{table}

In the simulation results as shown in Table \ref{table:result} , our method (RHS) shows noticeable advantages, especially in more complex terrain with severe roughness. In mild roughness terrain, where the predictions of the ISO probability curves closely align with our model, the improvements in success rate and search time are modest, with only a $9.52\%$ increase in success rate over ISO. However, as terrain complexity increases (e.g., in severe roughness conditions), the benefits of our dynamic partitioning method based on the smart agent-based model become more pronounced. RHS shows a significant $81.25\%$ improvement in success rate and a $48.63\%$ reduction in search time over ISO. When compared to TPS, which shares the same reference model but lacks dynamic partitioning, RHS consistently demonstrates better performance. Both RHS and TPS exhibit a tendency to locate the target within the first 90 minutes, as they prioritize high-probability areas to maximize the likelihood of success early in the mission. In contrast, ISO follows a more even coverage approach, distributing UAVs across probability curves, which may explain its lower success rate in later stages of the search as the area expands. This shows that our strategy, by focusing on high-probability regions early on, significantly enhances the chance of finding the target before the search area becomes too large. Simulation examples of the proposed search strategy's performance in the dynamic probability map with different terrain map can been in Fig. \ref{fig:results}.

\section{Conclusion} \label{conclusion}

In this work, we propose a solution for WiSAR using a multi-UAV system, focused on prioritizing areas with the high likelihood of finding the target. Our approach integrates terrain features and the lost person's profile, simulating self-propelled agents that move and interact with the environment from the LKP. This generates a dynamic probability density map reflecting the potential location of the lost person. Using this probability map, we formulate the multi-UAV search problem in a continuous domain, incorporating an approximation function to manage collision avoidance and communication between UAVs. To prevent UAVs from converging on the same high-probability areas and to encourage exploration, we apply dynamic task allocation using Voronoi partitioning, which divides the search area among UAVs. Our simulation results demonstrate that the proposed strategy significantly outperforms benchmark methods in terms of both mission success rate and search efficiency.

In future work, we aim to extend our search problem to three dimensions by incorporating UAV altitude changes, aligning more closely with real wilderness scenarios. In such environments, UAVs must adapt to significant terrain elevation changes to maintain a consistent FOV. Additionally, we plan to refine our search formulation by integrating detection probability (POD) variations based on complex terrain features. For instance, POD in densely forested areas is expected to be lower than in open environments like deserts. Lastly, we will implement our algorithm in real UAV teams to assess its performance in real-world search and rescue missions, providing valuable insights and validation of our approach.

% \usepackage{multirow}
% \usepackage{multirow}
% \usepackage{multirow}

% or


\ifCLASSOPTIONcaptionsoff
  \newpage
\fi

\bibliographystyle{IEEEtran}
\bibliography{mybib}



%\vfill


\end{document}